\documentclass[11pt]{article}
\usepackage{coling2020}
\usepackage{times}
\usepackage{wrapfig}
\usepackage{url}
\usepackage{latexsym}
\usepackage{color}
\usepackage{booktabs}
\usepackage{subfigure} 
\usepackage{multirow}
\usepackage{verbatim}
\usepackage{amsfonts}
\usepackage{float}
\usepackage{bm}
\usepackage{comment}
\usepackage{array}
\usepackage{graphicx}  
\usepackage{amsmath}
\usepackage{blindtext}
\usepackage{wrapfig}

\title{Learning to Decouple Relations: Few-Shot Relation Classification with Entity-Guided Attention and Confusion-Aware Training 
}
\author{Yingyao Wang$^1$\thanks{This work is done during the first author's internship in JD AI Research.} , 
  Junwei Bao$^2$,
  Guangyi Liu$^3$,
  Youzheng Wu$^2$,\\
  \bf Xiaodong He$^2$,
  Bowen Zhou$^2$,
  Tiejun Zhao$^1$ \\
  $^1$Harbin Institute of Technology, Harbin, China\\
  $^2$JD AI Research, Beijing, China\\
  $^3$The Chinese University of Hong Kong, Shenzhen, China\\
  \tt{\{baojunwei,wuyouzheng1,xiaodong.he,bowen.zhou\}@jd.com} \\
  \tt {yywang@hit-mtlab.net},{guangyiliu@link.cuhk.edu.cn},{tjzhao@hit.edu.cn} 
  }
\date{}

\begin{document}
\maketitle
\begin{abstract}
This paper aims to enhance the few-shot relation classification especially for sentences that jointly describe multiple relations.
Due to the fact that some relations usually keep high co-occurrence in the same context,
previous few-shot relation classifiers struggle to distinguish them with few annotated instances.
To alleviate the above \textit{relation confusion problem}, we propose CTEG, a model equipped with two mechanisms to learn to decouple these easily-confused relations.
On the one hand, an \textbf{E}ntity-\textbf{G}uided \textbf{A}ttention (EGA) mechanism, which leverages the syntactic relations and relative positions between each word and the specified entity pair, is introduced to guide the attention to filter out information causing confusion.
On the other hand, a \textbf{C}onfusion-\textbf{A}ware \textbf{T}raining (CAT) method is proposed to explicitly learn to distinguish relations by playing a pushing-away game between classifying a sentence into a true relation and its confusing relation.
Extensive experiments are conducted on the FewRel dataset, and the results show that our proposed model achieves comparable and even much better results to strong baselines in terms of accuracy.
Furthermore, the ablation test and case study verify the effectiveness of our proposed EGA and CAT, especially in addressing the relation confusion problem.
\end{abstract}

\section{Introduction}
Relation classification (RC) aims to identify the relation between two specified entities in a sentence. 
Previous supervised approaches on this task heavily depend on human-annotated data, which limit their performance on classifying the relations with insufficient instances.
Therefore, making the RC models capable of identifying relations with few training instances becomes a crucial challenge.
Inspired by the success of few-shot learning methods in the computer vision community\cite{Vinyals2016Matching} \cite{Sung1711,SantoroBBWL16}, \newcite{Han2018FewRel} first introduce the few-shot learning to RC task and propose the FewRel
dataset.
Recently, many works focus on this task and achieve remarkable performance \cite{gao2019hybrid,snell2017prototypical,YeMulti}. 

Previous few-shot relation classifiers perform well on sentences with only one relation of a single entity pair.
However, in real natural language, a sentence usually jointly describes multiple relations of different entity pairs.
Since these relations usually keep high co-occurrence in the same context, previous few-shot RC models struggle to distinguish them with few annotated instances.
For example, Table~\ref{sentence} shows three instances from the FewRel dataset, where each sentence describes multiple relations with corresponding keyphrases highlighted (colored) as evidence.
When specified two entities (bold black) in the sentence, there is a great opportunity for the instance to be incorrectly categorized as a {\color{red}{confusing relation}} (red) instead of the {\color{blue}{true relation}} (blue).
Specifically, 
the first instance should be categorized as the true relation `\textit{parents-child}' based on the given entity pair and natural language (NL) expression `\textit{a daughter of}'.
However, since it also includes the NL expression `\textit{his wife}', 
it is probably misclassified into this confusing relation `\textit{husband-wife}'.
In this paper, we name it as a \textbf{relation confusion problem}.



\begin{table}[t]
	\renewcommand\arraystretch{1}
	\small
	\centering
	\begin{tabular}{c|c|p{9.2cm}}

		\toprule
			\color{blue}{\textbf{True Relation}} & \color{red}{\textbf{Confusing Relation}}  &\textbf{Instance} \\ 
		\midrule
		
		\color{blue}{{parents-child}} & \color{red}{{husband-wife}}  &{She was {\color{blue}{a daughter of}} prince Wilhelm of Baden and {\color{red}{his wife}} \textbf{\emph{princess Maria of Lichtenberg}}, as well as an elder sister of \textbf{\emph{prince Maximilian}}.} \\ 
		\midrule
		
		\color{blue}{{husband-wife}} & \color{red}{{uncle-nephew}} & He was the youngest son of \textbf{\emph{Prescott Sheldon Bush}} and {\color{blue}{his wife}} \textbf{\emph{Dorothy Walker Bush}}, and {\color{red}{the uncle of}} former president George W Bush. \\

		\midrule
		{\color{blue}{{uncle-nephew}}} & \color{red}{{parents-child}}  & \textbf{\emph{Snowdon}} is {\color{red}{the son of}} princess Margaret, countess of Snowdon, and the 1st earl of Snowdon, thus he is {\color{blue}{the nephew of}} \textbf{\emph{queen Elizabeth ii}}.\\
		\bottomrule
	\end{tabular}
	\caption{Example sentences containing confusing relations.
		Their specified entities are marked as italics in bold. 
		The {\color{blue}{\textbf{blue}}} and {\color{red}{\textbf{red}}} words respectively correspond to true and confusing relations.}
	\label{sentence}
		\vspace{-0.5cm}
\end{table}
To address the relation confusion problem, it is crucial for a model to
be aware of which NL expressions cause confusion and learn to avoid mapping the instance into its easily-confused relation.
From these perspectives, we propose two assumptions. 
Firstly, \textit{in a sentence, words that keep high relevance to the given entities are more important in expressing the true relation}. Intuitively, the specified entity information is crucial to identify the true relations. 
Secondly, \textit{explicitly learning of mapping an instance into its confusing relation with augmented data in turn boosts a few-shot RC model on identifying the true relation}.
Based on these assumptions, we propose CTEG, a few-shot RC model with two novel mechanisms:
(1) An \textbf{E}ntity-\textbf{G}uided \textbf{A}ttention (EGA) encoder, which leverages the syntactic relations and relative positions between each word and the specified entity pair to
softly select important information of words expressing the true relation and filter out the information causing confusion.
(2) A \textbf{C}onfusion-\textbf{A}ware \textbf{T}raining (CAT) method, which explicitly learns to distinguish relations by playing a pushing-away game between classifying a sentence into a true relation and its confusing relation.
In addition, inspired by the success of pre-trained language models, our approaches are based on BERT \cite{devlin2018bert}, which has been proved effective especially for few-shot learning tasks.

Specifically, the backbone of the encoder of our model is a transformer equipped with the proposed \textbf{EGA} which guides the calculation of self-attention distributions by weighting the attention logits with entity-guided gates.
The gates are used to measure the relevance between each word and the given two entities.
Two types of information for each word are used to calculate its gate.
One is the \textit{relative position} \cite{Zeng2015DistantSF} information, which is the relative distance between a word and an entity in the input sequence.
The other is \textit{syntactic relation} which is proposed in this paper, defined as the dependency relations between each word and the entities.
Based on these information, the entity-guided gates in EGA are able to select those important words and control the contribution of each word in self-attention.

We also propose \textbf{CAT} to explicitly force the model to asynchronously learn the classification from an instance to its true relation and its confusing relation.
After each training step, the CAT first selects those misclassified sentences, and regards the relations they are misclassified into as the confusing relations.
After that, The CAT uses these misclassified instances and their confusing relations as augmented data to conduct an additional training process, which aims to learn the mapping between these instances into the confusing relations. 
Afterwards, the CAT adopts the KL divergence~\cite{10.2307/2236703} to teach the model to distinguish the difference between the true and confusing relations, which benefits the true relation classification from the confusing relation identification.


The contributions of this paper are summarized as follows: 
(1) We propose an Entity-Guided Attention encoder, which can select crucial words and filter out NL expressions causing confusion based on their relevance to the specified entities.  
(2) We propose a Confusion-Aware Training process to enhance the model with the ability of distinguishing true and confusing relations. 
(3) We conduct extensive experiments on few-shot RC dataset FewRel, ans the results show that our model achieves comparable and even much better results to strong baselines.
Furthermore, ablation and case studies verify the effectiveness of the proposed EGA and CAT, especially in addressing the relation confusion problem.

\section{Methodology}

\begin{figure}[t]
\centering
\includegraphics[width=15cm]{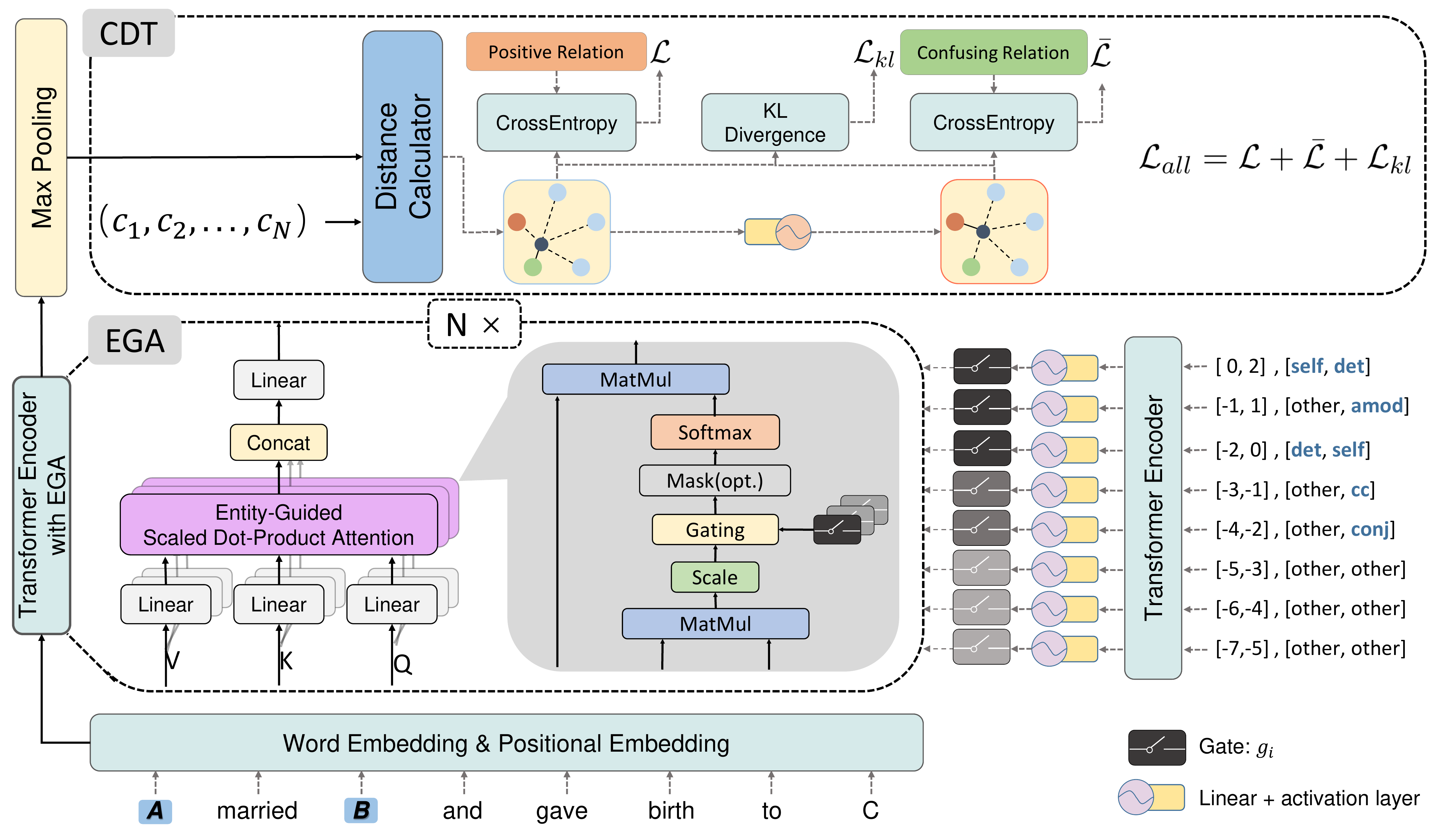}
\caption{The framework of our model CTEG including  \textbf{E}ntity-\textbf{G}uided \textbf{A}ttention (EGA) and \textbf{C}onfusion \textbf{A}ware \textbf{T}raining (CAT) mechanisms.}
\label{model}
\end{figure} 

\subsection{EGA: Entity-Guided Attention Encoder}
The inputs of our model include a sentence $S=w_{1},...,w_{n}$ with $n$ words, and two pairs of integers $s_1 =(l_1,r_1)$ and $s_2=(l_2,r_2)$ representing the start and end positions of the two specified entities. Firstly, we convert the words into a sequence of vectors $\bm e_{1}^{w},...,\bm e_{n}^{w}$, using an embedding layer initialized by BERT. We then use two types of relevance information, i.e., relative position and syntactic relation, between each word and the specified entity pair to construct entity-guided gates for information selection.
\paragraph{Relative Position.}
Relative position information is typically used in relation classification task~\cite{Zeng2015DistantSF}, which is defined as the relative distances $pos_1$ and $pos_2$ from the current word to the two specified entities in the sentence. 
The relative position information of the $i$-th word is represented as $\bm e^{pos}_i = [\bm e_{i}^{pos1}, \bm e_{i}^{pos2}]$, where $\bm e_{i}^{pos1},\bm e_{i}^{pos2} \in \mathbb{R}^{d_{pos}}$ are the embeddings of $pos_1$ and $pos_2$.

\begin{figure*}[t]
	\centering
	\subfigure[]{\label{dependency tree:a}\includegraphics[width=0.49\textwidth]{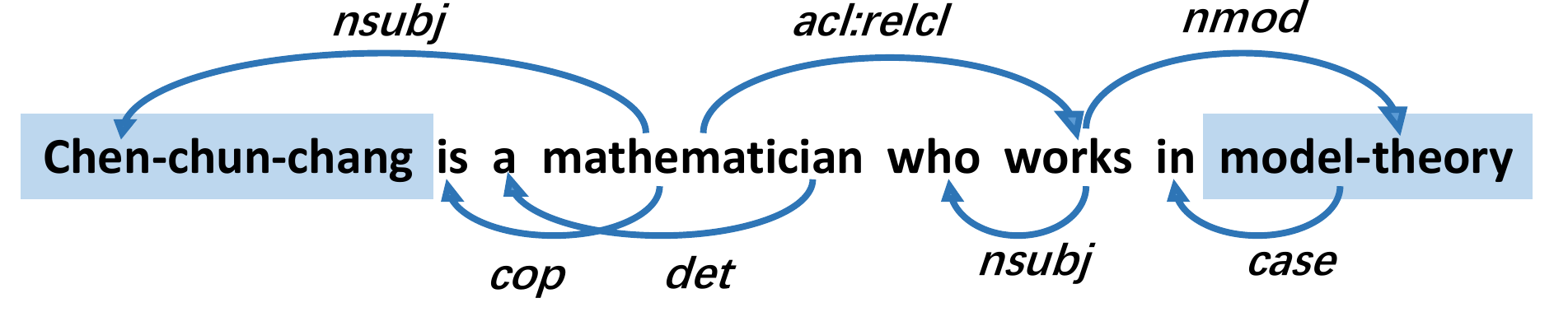}}
	\subfigure[]{\label{dependency tree:b}\includegraphics[width=0.49\textwidth]{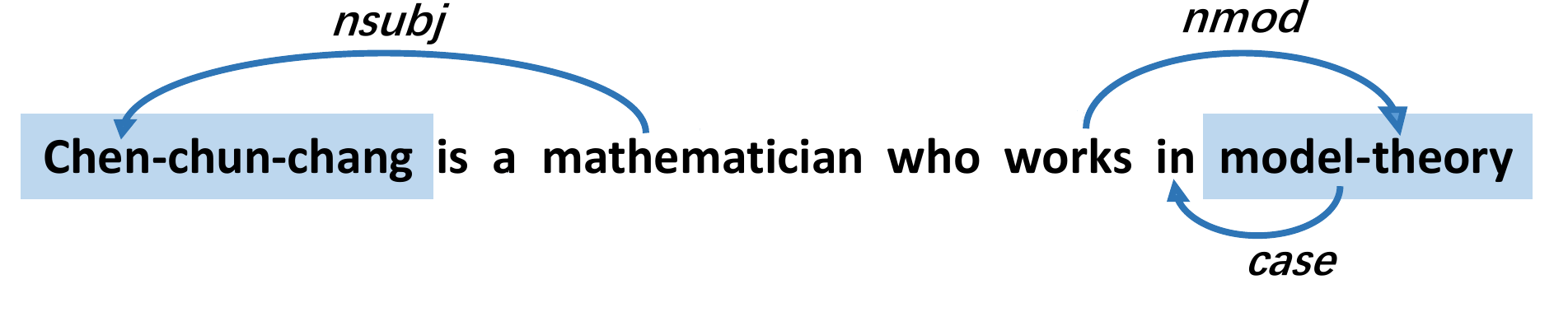}}
	\caption{The dependency tree of a sentence and the paths used as syntactic relations in EGA. }
	\label{dependency tree}
	\vspace{-0.5cm}
\end{figure*}

\paragraph{Syntactic Relation}
Except for the relative position, we further introduce the syntactic relations to measure the relevance between each word and the specified entities.
The syntactic relations are derived based on dependency parse trees, which are obtained from the Standford Parser\footnote{\url{https://nlp.stanford.edu/software/lex-parser.shtml}}. 
For example, Figure~\ref{dependency tree:a} shows the original dependency tree of the sentence \textit{``Chen-chun-chang is a mathematician who works in model-theory"}, where \textit{``Chen-chun-chang"} and \textit{``model-theory"} are entities.
In this paper, we assume that words that directly connect to the given entities are more important in expressing the true relations.
Therefore, dependency relations that connect the specified entities and other words are remained and the others are discarded, which derives a pruned dependency tree, as one shown in Figure~\ref{dependency tree:b}.
Based on the pruned dependency tree, each word in the sentence is assigned two tags $t_i=(t_{i,1},t_{i,2})$ as the proposed \textbf{syntactic relations}.
Taking the tag $t_{i,1}$ of word $w_i$ which corresponds to the first entity as an example, if $w_i$ is part of the first entity, the tag $t_{i,1}$ is assigned the value `\textit{self}', and if $w_i$ is directly connected to the first entity in the dependency tree, $t_{i,1}$ is assigned the dependency relation, e.g., `\textit{nmod}'.
In addition, if $w_i$ is neither connected to nor part of the first entity, $t_{i,1}$ is assigned `\textit{other}'. 
Based on the above strategy, the syntactic relations of the sentence in Figure~\ref{dependency tree} are shown in Table~\ref{syntactic pos}.
Finally, the two dependency tags of each word $t_i=(t_{i,1},t_{i,2})$ are converted into continuous vectors based on an embedding lookup operation, and then concatenated into a vector $\bm e^{syn}_i = [\bm e_{i}^{syn1}, \bm e_{i}^{syn2}]$, where $\bm e_{i}^{syn1},\bm e_{i}^{syn2} \in \mathbb{R}^{d_{syn}}$. 
\begin{table}[t]
	\centering
	\small
	\begin{tabular}{c|cccccccc}
		\toprule
		\bf Words &\bf \textit{chen chun chang} &  \textit{is} & \textit{a} & \textit{mathematician} & \textit{who} & \textit{works} & \textit{in} &\bf \textit{model theory}  \\
		\midrule
		\bf $t_{i,1}$ &self & other &other &nsubj &other &other &other &other \\
		\bf $t_{i,2}$ &other & other &other &other &other &nmod &case&self \\
		\bottomrule
	\end{tabular}
	\caption{The syntactic relations corresponding to each word of the given sentence. }
	\label{syntactic pos}
		\vspace{-0.5cm}
\end{table}
\paragraph{Entity-Guided Gate}
The proposed EGA learns entity-guided gates $\bm G = (g_1,...,g_n)$ for all words in a given sentence based on the above two types of information.
Intuitively, if a word $w_i$ is directly connected to the given entities in the dependency tree, the corresponding information tends to be more important.
Specifically, the relative position embedding and the syntactic relation embedding are first concatenated into $\bm e_i^p=[\bm e_i^{pos}, \bm e_i^{syn}]$, where $\bm e_i^p \in \mathbb{R}^{2d_{pos}+2d_{syn}}$.
We then adopt a transformer encoder \cite{46201} followed by a single layer feed-forward neural network (FNN) with $\mathtt{sigmoid}(\cdot)$ activation to derive a sequence of entity-guided gates as follows:
\begin{equation}
\bm (\bm h^p_{1},...,\bm h^p_{i},...,\bm h^p_{n}) = \mathtt{TransEnc}(\bm e^p_{1},...,\bm e^p_{i},...,\bm e^p_{n})
\end{equation}
\begin{equation}
g_i = \mathtt{sigmoid}(\bm W^g \bm h_i^p + \bm b^g)
\end{equation}

\paragraph{Gated Self-Attention}
A pre-trained transformer encoder with $M$ layers equipped with the proposed EGA is used to learn the representation for a sentence.
The backbone of each layer is a self-attention layer, which calculates attention weights for word pairs in the sentence.
We define self-attention weights of the $t$th-layer as $\bm{Att}^{t}$, and the corresponding hidden states of the sentence is represented as $\bm{H}^t$. 
To obtain the attention weights $\bm{Att}^{t}$, the scaled attention scores $\frac{\bm Q^t {\bm K^t}^\top}{\sqrt{d_k}}$ is multiplied by the entity-guided gate $\bm G$ with broadcasting followed by a $\mathtt{softmax(\cdot)}$ operation.
The gated self-attention and the calculation of $\bm{H}^t$ are formalized as follows, where $\bm W_k$,$\bm W_q$,$\bm W_v$ are trainable parameters.
\begin{equation}
(\bm Q^t,\bm K^t,\bm V^t) = (\bm {W_q},\bm {W_k},\bm {W_v})\bm H^{t-1}
\end{equation}
\begin{equation}
\bm{Att^t} = \mathtt{softmax}(
     \frac{\bm Q^t {\bm K^t}^\top \otimes \bm G}
     {\sqrt{d_k}})
\end{equation}
\begin{equation}
\bm{H}^t =\bm{Att}^{t}\bm{V}^{t}
\end{equation} 
Finally, vector $\bm s$ as the representation of the sentence is obtained based on Equation~\ref{Eqa6}, where $\bm H^M$ represents the output of the $M$-th layer of the encoder.
\begin{equation}  
\bm s = \mathtt{maxpooling}(\bm H^M ) \label{Eqa6}
\end{equation}

\subsection{Classification}
The classifier performs $N$-way-$K$-shot classification following few-shot learning paradigm and the prototypical network \cite{snell2017prototypical}.
Specifically, for a relation $r_j$ where $j \in [1,N]$, $K$ sentences are sampled from its instances firstly, and then these sentences are used to calculate the representation named prototype $\bm c_j$ of the relation. We define the representations of the $K$ sentences as ${\bm s_{j,1}^{c},...,\bm s_{j,K}}^{c}$, and prototype $\bm c_j$ is calculated as follows:
\begin{equation}
\bm c_j=\frac{1}{K}\sum_{k=1}^{K}\bm s_{j,k}^{c}
\end{equation}
Given the representation $\bm s^{q}$ of a sentence as query and the prototypes $(\bm c_1,...,\bm c_N)$ of $N$ relations, the model aims to classify $\bm s^{q}$ into one of the $N$ candidate relations.
We first obtain the distance distribution $\bm \delta=(\delta_{1},...,\delta_{N})$ by calculating the Euclidean Distance between $\bm s^q$ and each prototype.
Then, according to $\bm \delta$, the sentence will be classified into the nearest relation $\hat{r}$.
\begin{equation}
\bm{\delta} =({\left \|\bm s^{q}-\bm c_1 \right\|^2,...,\left \|\bm s^{q}-\bm c_N \right\|^2})
\end{equation}
\begin{equation}
\hat{r}=\arg\min_j{(\bm{\delta})}
\end{equation}
To enable the classifier to learn confusing relations, we further project the distance distrubution $\bm \delta$ into $\bm {\Bar{\delta}}$ via a FFN with a $\mathtt{tanh}(\cdot)$ activation function defined as follows.
The $\bm {\Bar{\delta}}$ is used to predict the confusing relation defined as $\Bar{r}$ during a confusion-aware training (CAT) stage which is introduced in Section~\ref{CAT}.
\begin{equation}
\bm {\Bar{\delta}} = \mathtt{tanh}(\bm W^{c} \bm \delta + \bm b^{c})
\end{equation}

\subsection{CAT: Confusion-Aware Training \label{CAT}}
The confusion-aware training is based on two asynchronous processes: \textit{true relation identification} and \textit{confusing relation identification}. During classifying a sentence, the former uses its true relation as the target, and the latter uses its confusing relation as the target. Specifically, given a sentence with its true relation as $r$, the training objective of the true relation identification is defined as:
\begin{equation}
\mathcal{L} = \mathtt{CrossEntropy}(\mathtt{OneHot}(r),\mathtt{Softmax}(\bm \delta))
\end{equation}
For the \textit{confusing relation identification}, we first pick up those misclassified sentences after each training step of true relation identification, and use their prediction results as the targets. In formulation, assuming the sentence is misclassified into an incorrect $\Bar{r}$, the objective function of the confusing relation identification $\Bar{\mathcal{L}}$ is defined as:
\begin{equation}
\Bar{\mathcal{L}} = \mathtt{CrossEntropy}(\mathtt{OneHot}(\Bar{r}),\mathtt{Softmax}(\bm{ \Bar{\delta}}))
\end{equation}
Besides, the KL divergence is adopted as another objective function, which allows the model to learn to perform confusion decoupling. The KL divergence has the ability to push away the distance distribution $\bm \delta$ and $\bm {\Bar{\delta}}$, and the formula is defined as follows:
\begin{equation}
\mathcal{L}_{kl} =-\mathtt{KL}( \mathtt{Softmax}(\bm \delta),\mathtt{Softmax}(\bm {\Bar{\delta}}))
\end{equation}
Through minimizing $\mathcal{L}_{kl}$, the model is able to explicitly learn to distinguish relations by playing a pushing-away game between classifying a sentence into a true relation and its confusing relation.
In other words, our model learns to explicitly decouple $r$ and $\Bar r$ for classification based on specified entities in a given sentence.
It is worth noting that, only those misclassified sentences are used for updating the objective $\mathcal{L}_{kl}$.
The final objective function of our model $\mathcal{L}_{all}$ is defined as $\mathcal{L}_{all} = \mathcal{L}+\Bar{\mathcal{L}}+\mathcal{L}_{kl}$.

\section{Experiments}
In this section, we report our experiment results from the following four aspects.
We first show the comparison results of our model CTEG and baselines on FewRel dataset in Section \ref{Results}.
We then demonstrate the effectiveness of the proposed entity-gated attention (EGA) and confusion-aware training (CAT) through the ablation studies in Section \ref{secAbl}.
In order to more intuitively and clearly show the role of EGA and CAT, we show their visualized examples in case study in Section \ref{case}.
Furthermore, we verify that our model is capable of addressing the \textit{relation confusion problem} to some extent in Section \ref{confusion}.

\subsection{Implementation Details}
\paragraph{Dataset} 
The FewRel dataset~\cite{Han2018FewRel} contains 100 relations, which are split up into $64$ for training, $16$ for validation and $20$ for testing. 
Each relation has 700 instances generated by distant supervision \cite{mintz2009distant}. 
All the instances are annotated with a  specified entity pair. 
\paragraph{Settings}

The dimension of word embedding is set to $768$ for consistency with the base model of BERT~\cite{devlin2018bert}.
The max length of the input is set to $100$.
Following BERT, the layer number $M$ of the transformer encoder with EGA is $12$, and all parameters in it is initialized with the pretrained BERT model.
The relative position and syntactic relation embedding dimensions are both set to $50$, and the transformer encoder for obtaining entity-guided gates is set up with hidden size as $230$, head number of self-attention as $2$.
In addition, the model is optimized by Adam algorithm with the learning rate and the weight decay as $1\times 10^{-5}$ and $1\times 10^{-6}$, respectively.  
\subsection{Baselines}
We implement {{four}} baselines on FewRel dataset: \textbf{Proto}, \textbf{Proto-HATT} \cite{gao2019hybrid}, \textbf{MLMAN} \cite{YeMulti} and \textbf{BERT-PAIR} \cite{gao-etal-2019-fewrel}.
All the baselines are based on the few-shot learning framework.
%
Specifically, for each training step, $N$ relations are first sampled from the training set.
For each of the above relation, $K$ out of 700 instances are sampled to construct a supporting set, based on which a relation representation named \textit{prototype} is calculated.
Given an instance of the $N$ relations to be classified, the models classify it by calculating the distances from it to $N$ prototypes.
\paragraph{Proto \& Proto-HATT} Both of the two models adopt the convolutional neural network (CNN) as encoders.
	{Proto} calculates the prototype by averaging the representations of the $K$-instances in the supporting set, and classify the query using the Euclidean Distance.
    Differently, {Proto-HATT} further proposes a hybrid attention scheme which includes an instance-level attention and a feature-level attention, where the former is used to highlight the crucial support sentences in calculating the prototype, and the latter is to select more efficient features when calculating distances. 
\paragraph{MLMAN} Different from the Proto and Proto-HATT, {MLMAN} encodes each query and the supporting set in an interactive way by considering their matching information on multiple levels.
	At local level, the representations of an instance and a supporting set are matched following the sentence matching framework \cite{chen-etal-2017-enhanced} and aggregated by max and average pooling.
	At instance level, the matching degree is first calculated via a multi-layer perception (MLP).
	Then, taking the matching degrees as weights, the instances in a supporting set are aggregated to obtain the class prototype for final classification.
\paragraph{BERT-PAIR} This model is based on the sentence classification model in BERT. The sentence to be classified is first paired with all the supporting instances, and then each pair is concatenated to a sequence. BERT takes this sequence as input and returns a relevance score, which is used to measure whether the given sentence expresses the same relation with the corresponding supporting instance.

\begin{table}[t]
     \small
	\centering
   
	\begin{tabular}{l|llll}
		\toprule
		\bf Model & \bf 5-way-1-shot & \bf 5-way-5-shot & \bf 10-way-1-shot & \bf 10-way-5-shot \\
		\midrule
		\midrule
		Proto \cite{Han2018FewRel}& $72.65$ / $74.52$ & $86.15$ / $88.40$ & $60.13$ / $62.38$ & $76.20$ / $80.45$ \\
		Proto-HATT \cite{gao2019hybrid}& $75.01$ / $--$ & $87.09$ / $90.12$ & $62.48$  / $--$& $77.50$ / $83.05$ \\
		MLMAN \cite{YeMulti}& $78.85$ / $82.98$ &$88.32$ / $92.66$ &$67.54$ / $73.59$ & $ 79.44$ / $87.29$\\
		BERT-PAIR \cite{gao-etal-2019-fewrel} & $\textbf{85.66}$ / $\textbf{88.32}$ &$89.48$ / $93.22$ &$\textbf{76.84}$ / $80.63$ & ${81.76}$ / $87.02$\\
	
		{CTEG (This work)} & ${84.72}$ / $88.11$ & $\textbf{92.52}$ / $\textbf{95.25}$ & ${76.01}$ / $\textbf{81.29}$& $\textbf{84.89}$ / $\textbf{91.33}$ \\
		\midrule
		 ~~w/o CAT &  $83.79$ & $91.71$ & $74.25$ & $84.46$ \\
		 ~~w/o EGA &  $72.94$ & $86.71$ & $61.88$ & $77.03$ \\
		 ~~~~w/ Pos &  $82.31$ & $91.44$ & $72.41$ & $83.98$ \\
		 ~~~~w/ Syn &  $83.61$ & $91.78$ & $73.94$ & $84.89$ \\

		\bottomrule
	\end{tabular}
	\caption{The main classification accuracy of baselines and our model are shown in \textbf{Validation / Test} format, where the test results are from FewRel public leaderboard\footnote{https://github.com/thunlp/FewRel}. All ablation results reported in this section are on the validation set.}
	\label{all}
	\vspace{-0.5cm}
\end{table}

\subsection{Comparison with Baselines \label{Results}}
Same as Proto, we set $N=5,10$ and $K=1,5$ for $N$v$K$ few-shot learning.
Average accuracy is used as the evaluation metric to evaluate the relation classification performance.
The results in Table~\ref{all} show that our model EGA with CAT, named \textbf{CTEG}, outperforms the three strong baselines including Proto, Proto-HATT and MLMAN by a significant margin on all the settings.
These improvements are mainly brought by our EGA and CAT, which help the model to classify those easily-confused instances into correct ones.
In addition, applying pre-trained BERT also contributes to improving the performance.
Compared with BERT-Pair, our CTEG achieves better result on 5v5, 10v1 and 10v5 settings and comparable results on 5v1 settings on the test set, while on the dev set our CTEG is slight lower than BERT-pair on 5v1 and 10v1 settings.
We think that the lower performance on the dev set on 5v1 and 10v1 is due to the fact that BERT-Pair encodes two sentences together which benefits for information fusion, while models based on prototypical network rely on larger $K$ supporting facts to get a better prototype.

\subsection{Ablation Study\label{secAbl}}
We conduct ablation study and show the results in Table~\ref{all}.
Firstly, we turn off the CAT of our full model, which is represented as ``\textbf{w/o CAT}''.
In this case, the average results drops $0.43$-$1.76$ point on the four settings.
These drops indicate that the CAT has the ability to improve the classification performance.
We then report three groups of results to verify the effectiveness of EGA.
Specially, our model without EGA which only adopts the BERT as the encoder is denoted as ``\textbf{w/o EGA}''.
It is worth noting that in this case, the model can not identify which words in a given sentence are entities.
When the EGA is removed, the performance decreases obviously by $5.81$-$14.13$ point.
It is proved that the entity information is crucial for relation classification.
Furthermore, ``\textbf{w/ Pos}'' means the entity-guided gates in EGA are calculated only using the relative position information, and ``\textbf{w/ Syn}'' only using the syntactic relation information. Compared with ``\textbf{w/o EGA}'', the results of these two groups are significantly improved.
It shows that the syntactic relation information is more powerful than the relative position information, which means considering the dependency relations between each word and the specified entity pair boosts the performance of simply adopting traditional relative position information.
In addition, it can be seen that the smaller size of the supporting set (1-shot v.s. 5-shot), the more absolute gain our CAT and EGA modules achieve.
This phenomenon shows that our method performs well with fewer available supporting instances.
\begin{table}[h]
  \centering
  \small
  \begin{tabular}{l|c|c|c|c}
		\toprule
         \bf Model   & CTEG w/ Syn & QGG & CTEG & FHG\\
        \midrule
		\midrule
        \bf 5-way-5-shot   & $91.78\pm0.22$ & $90.64 \pm 0.28$ & $92.52 \pm 0.31$ &  $90.57 \pm 0.11$\\
        \bottomrule

  \end{tabular}
  \caption{Ablation Results on How, What, and When to Gate.}
  \label{ablation}
  	\vspace{-0.5cm}
\end{table}
\paragraph{How to Gate}
The self-attention mechanism is used to update the representation of each \textit{query} word by fusing the information of all \textit{key} words in a given sentence.
In this process, an attention score is calculated to leverage the contribution of each key word.
In this paper, we propose to use gates to further adjust these attention scores.
In our proposed EGA, each \textit{entity-guided gate} reflects the relation between the key word and the specified entity pair, which is different for all key words.
We also implement a baseline \textbf{QGG} with \textit{query-guided gates}, where each gate reflects the relation between the key word and the query word.
Specifically, the relation is modeled based on their syntactic relation if the key word is a specified entity, otherwise a `\textit{other}' relation.
The results of using these two kinds of gates in Table~\ref{ablation} shows that our model \textbf{CTEG w/ Syn} only modeling syntactic relations outperforms the \textbf{QGG} baseline, which further verifies that our EGA with entity-guided gates has the ability of effectively leveraging specified entity information to select input information.

\paragraph{What and When to Gate}
In our EGA, the entity-guided gates are used since the beginning of the encoding process by multiplying them with the self-attention scores in each transformer layer.
It means that the information of the words is selected during learning the representation of them.
Another baseline is to multiply gates with the \textbf{F}inal transformer \textbf{H}idden states of the words as \textbf{G}ating mechanism, which is defined as \textbf{FHG}.
In this case, the information of all words has been fully fused before adopting gating mechanism for selection.
As the results shown in Table~\ref{ablation}, compared with our model \textbf{CTEG}, the accuracy of the \textbf{FHG} drops $1.95$ point.
The results indicate that earlier to gate the attention score during encoding is more reasonable than only to adopt gating at the final hidden states, which verifies the effectiveness of the proposed EGA.

\subsection{Case Study\label{case}}
\begin{figure}[h]
	\centering
	\includegraphics[width=0.6\textwidth]{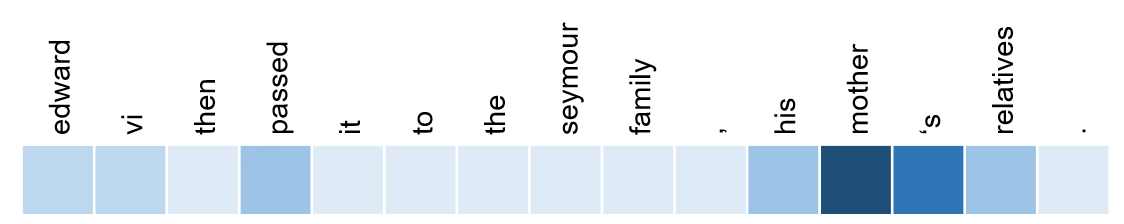}
	\caption{An example of the entity-guided gates of a given sentence.}
	\label{fig:my_label}
		\vspace{-0.5cm}
\end{figure}

\begin{figure}[h]
    \centering
	\includegraphics[width=12cm]{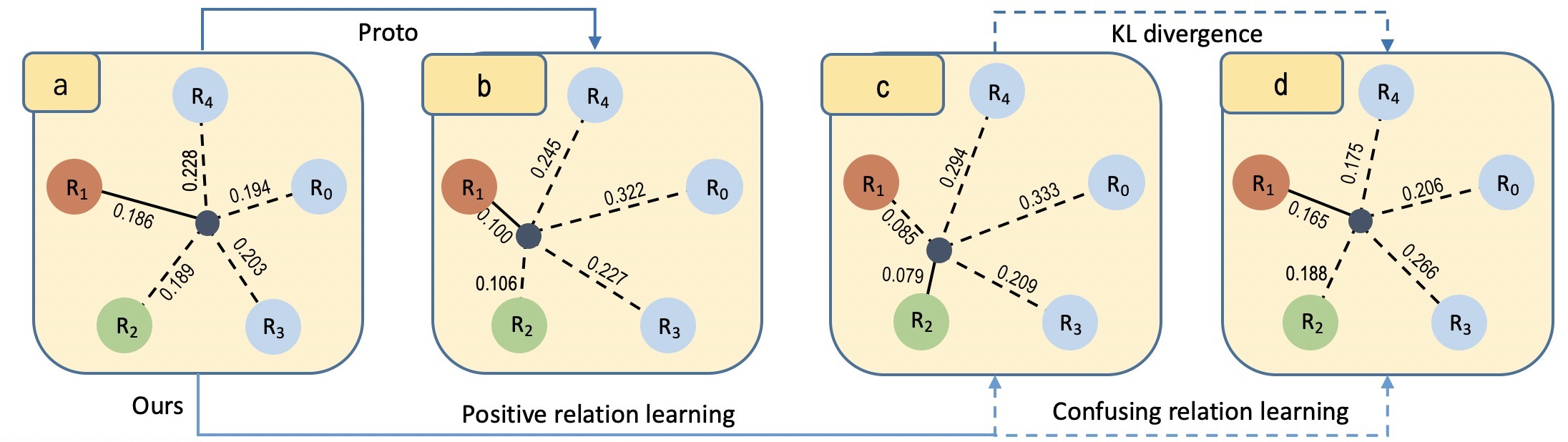}
	\caption{Distances between a given sentence and five candidate relations calculated by different models. where (a) is from a random initialized network, (b) is from the Proto network, (c) and (d) are respectively from our \textit{true relation training} and \textit{confusing relation training}.}
	\label{fig:distance}
	\vspace{-0.5cm}
\end{figure}

\paragraph{EGA visualized example} The entity-guided gates in EGA are expected to emphasize the words which are more related to the true relation. To verify the effectiveness of EGA intuitively, we show the entity-guided gates heat map of a given instance in Figure~\ref{fig:my_label}.
This instance is sampled from  '\textit{parent-children}' relation in the validation set of FewRel. 
As shown in the map, the words '\textit{his mother is}' are given higher scores.
Obviously, the three words are important for expressing the '\textit{parent-children}' relation.  
\paragraph{CAT visualized example} In Figure~\ref{fig:distance}, we visualize the distance distributions between the given sentence and its candidate relations. The four subfigures respectively show the distance distributions calculated by different models including our \textit{true relation identification} and \textit{confusing relation identification}.
Among the five candidate relations, $R2$ in green is the true relation of the sentence, and $R_1$ in red is the confusing relation that the sentence is usually misclassified into. 
Each edge in the subfigures represents the distance from the sentence to the corresponding relation, and the solid edge indicates the nearest one. Specifically, (a) is the distances calculated by a randomly initialized network. (b) is the classification result of Proto, in this case, the query is misclassified into $R_1$. (c) and (d) are the final classification results of our CAT. The distance distribution between the query and the confusing relation calculated by our CAT is shown in (d), and it can be seen that the model succeeds in making the query closer to the confusing relation $R_2$ as we expected. After that, the distance distribution information is propagated to the true relation training by KL divergence, this operation is used to push the distance distribution of the true relation prediction away from the distribution of the confusing relation. As (c) shows, the sentence is pushed away from $R_1$ and get closer to the true relation $R_2$. This example validates our assumption that explicit learning of confusing relations facilitates the identification of true relations.

\subsection{Relation Confusion Problem\label{confusion}}
\begin{figure*}[t]
\centering
	\includegraphics[width=16cm]{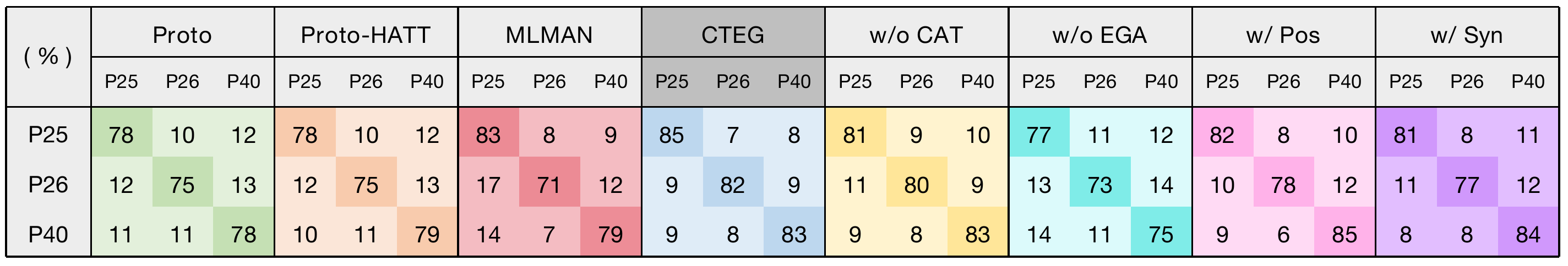}
	\caption{Confusion matrices of the three easily confused relations, where different colors represent the classification results of different models.}
	\label{cm}
	\vspace{-0.5cm}
\end{figure*}

In this section, we discuss the effectiveness of our model on confusion decoupling. In order to intuitively show the classification performance on the confusing relations, we use the confusion matrices as our evaluation metric.
\paragraph{Confusing Relations Selection} 
We first analyze the classification results of the baseline models Proto and Proto-HATT. Based on our statistics, we find three of the 16 relations in the FewRel validation set that are most easily confused with each other. Their relation indexes are \textbf{\textit{P25}}, \textbf{\textit{P26}} and \textbf{\textit{P40}}, and the corresponding true relations are ``\textbf{\textit{Parents-Child}}'',``\textbf{\textit{Husband-Wife}}'' and ``\textbf{\textit{Uncle-Nephew}}''. We test our model and the baseline models under the 5-way-5-shot configuration. For the three easily-confused relations, we respectively record the number of their sentences which are correctly classified and misclassified into the other two relations, and use the results to conduct the confusing matrices. 
\paragraph{Improvement of Relation Confusion Problem} 
As shown in Figure~\ref{cm}, we report the classification results of different models on the three confusing relations \textit{P25}, \textit{P26} and \textit{P40}.
In the confusion matrices, the horizontal axis represent the true relation of the sentences, and the vertical axis represent the classification results of these sentences by different models.
For each matrix, supposing a given relation such as \textit{P25} has $X$ sentences to participate in the test, and the numbers of the sentences classified into \textit{P25}, \textit{P26} and \textit{P40} are respectively $a$,$b$ and $c$, than the elements in the first row of the matrix are calculated as $(a,b,c/X)$. 
Given a relation, we expect the models classify more its sentences into the true relation, and fewer its sentences into confusing relations.
From this perspective, through comparing the confusion matrices of ``CTEG'' and the baseline models, it can be seen that our full model CTEG achieves the best performance in identifying these easily confused relations.
``w/o EGA'' has the weakest ability to decouple the confusing relations, because it is not provided with any entity information to identify the true relation.
Based on results of ``w/o EGA'', ``w/ Pos'' and ``w/ Syn'' we can see that both of the relative position and the syntax position bring significant improvements.
In addition, compared with our full model, the performance of ``w/o CAT'' proves that the CAT help to decouple the confusing relations.

\section{Related Work}
\paragraph{Few-shot Relation Classification} Relation classification (RC) aims to identify the semantic relation between two entities in a sentence.
It is an important task in natural language processing community and has attracted more and more attention over past few years. \cite{jia-etal-2019-arnor,feng2018reinforcement,le2018large,adel2017global,yang2016position}.
Previous supervised approaches on this task heavily rely on labeled data for training, that limits their ability to classify the relations with insufficient instances.
To address this problem, \newcite{Han2018FewRel} first introduce few-shot learning to RC task.
The few-shot learning paradigm has been proved effective in the computer vision community and has many applications~\cite{Vinyals2016Matching,Sung1711,SantoroBBWL16}.
Earlier works on few-shot RC are based on the widely used few-shot learning model prototypical network~\cite{snell2017prototypical,YeMulti}.
Recently, the pre-trained language models (LM) has shown significant power in many natural language processing tasks.
To this end, \newcite{gao2019fewrel} adopt the most representative pre-trained LM BERT \cite{devlin2018bert} to few-shot RC, and their work shows that BERT brings significant improvements on classification performance.
Furthermore, the approach proposed by \newcite{Livio2019Matching} are also based on BERT and achieve the state-of-art result on the few-shot RC task.

\paragraph{Syntactic Relation} Previous RC models usually use the relative position information to identify which words are the entities in a sentence, e.g., \newcite{zeng2015distant}.
In addition, the syntax information of the sentences is proved useful in many natural language processing tasks \cite{falenska2019non,ma2020entity,Chen2017TranslationPW}.
Inspired by \newcite{yang-etal-2016-position}, which adopt the dependency parse tree for RC \cite{ma2020entity}, we also introduce the dependency relation as another type of position to emphasize the specific entities, and propose a novel application of the syntax positions. 

\section{Conclusions}
In this paper, we propose CTEG equipped with two novel mechanisms, namely the Entity-Guided Attention (EGA) and the Confusion-Aware Training (CAT), to address the relation confusion problem in few-shot relation classification (RC).
We conduct extensive experiments on benchmark dataset FewRel, and experiment results shows that our model achieves significant improvements on few-shot RC.
Ablation studies verify the effectiveness of the proposed EGA and CAT mechanisms.
Case study and further analysis demonstrate that our model has the ability of decoupling easily-confused relations.
\section*{Acknowledgments}
We are grateful to the anonymous reviewers. The work of this paper is funded by the project of National High Technology Research and Development Program of China (No. 2018YFC0830700).

\bibliography{coling2020}
\bibliographystyle{coling2020}
\end{document}